\documentclass[letterpaper, 10 pt, conference]{ieeetran}  % Comment this line
\usepackage{graphicx} % for pdf, bitmapped graphics files
\usepackage{amsmath} % assumes amsmath package installed
\usepackage{amssymb}  % assumes amsmath package installed
\usepackage{url}

\usepackage[utf8]{inputenc}

\usepackage[firstpage]{draftwatermark}
\SetWatermarkText{
\begin{minipage}{15cm}\centering
\vspace{-25cm}
\textit{To appear in The International Conference on Information Fusion (FUSION) 2016}
\end{minipage}
}
\SetWatermarkScale{1}
\SetWatermarkLightness{0}
\SetWatermarkFontSize{10pt}
\SetWatermarkAngle{0}

\def\b#1{\boldsymbol{#1}}

\title{\LARGE \bf An Open Source, Fiducial Based, Visual-Inertial\\ Motion Capture System}

\author{
\IEEEauthorblockN{Michael Neunert}
\IEEEauthorblockA{Agile \& Dexterous Robotics Lab\\
ETH Zürich, Switzerland\\
Email: neunertm@ethz.ch}
\and
\IEEEauthorblockN{Michael Bloesch}
\IEEEauthorblockA{Autonomous Systems Lab\\
ETH Zürich, Switzerland\\
Email: bloeschm@ethz.ch}
\and
\IEEEauthorblockN{Jonas Buchli}
\IEEEauthorblockA{Agile \& Dexterous Robotics Lab\\
ETH Zürich, Switzerland\\
Email: buchlij@ethz.ch}
}

\begin{document}

\maketitle
\thispagestyle{empty}
\pagestyle{empty}

%%%%%%%%%%%%%%%%%%%%%%%%%%%%%%%%%%%%%%%%%%%%%%%%%%%%%%%%%%%%%%%%%%%%%%%%%%%%%%%%
\begin{abstract}
Many robotic tasks rely on the accurate localization of moving objects within a given workspace. This information about the objects' poses and velocities are used for control, motion planning, navigation, interaction with the environment or verification. Often motion capture systems are used to obtain such a state estimate. However, these systems are often costly, limited in workspace size and not suitable for outdoor usage. Therefore, we propose a lightweight and easy to use, visual-inertial Simultaneous Localization and Mapping approach that leverages cost-efficient, paper printable artificial landmarks, so called fiducials. Results show that by fusing visual and inertial data, the system provides accurate estimates and is robust against fast motions and changing lighting conditions. Tight integration of the estimation of sensor and fiducial pose as well as extrinsics ensures accuracy, map consistency and avoids the requirement for precalibration. By providing an open source implementation and various datasets, partially with ground truth information, we enable community members to run, test, modify and extend the system either using these datasets or directly running the system on their own robotic setups.
\end{abstract}

%%%%%%%%%%%%%%%%%%%%%%%%%%%%%%%%%%%%%%%%%%%%%%%%%%%%%%%%%%%%%%%%%%%%%%%%%%%%%%%%
\section{Introduction}
\subsection{Motivation}
For many tasks in mobile robotics, it is important to estimate a robot's state
with respect to its workspace, i.e. its pose and velocities expressed in an inertial coordinate system aligned to the robot's workspace. Such tasks include navigation, motion planning or manipulation.
One way to measure the position and orientation of a robot is to use a motion
capture system (such as e.g. Vicon, Optitrack or PTI Visualeyez). These systems are usually highly accurate and provide pose estimates w.r.t. to a calibrated reference system. However, these systems can be very costly and the user might be limited to a certain workspace size. Furthermore, many common systems such as Vicon and Optitrack use passive markers that reflect infrared light which limits their usage to indoor setups. Also, most systems require a tedious calibration procedure that needs to be repeated frequently to maintain accuracy. Since these systems do not have access to inertial data, they have to rely on finite differences of position measurements to estimate velocity information, which usually leads to highly quantized (compare Figure \ref{fig:rotVel}) or delayed data.

Another approach to state estimation is Visual Odometry (VO), which is sometimes also fused with inertial data. VO can provide very accurate local estimates of the robot motion. However, usually only a finite number of previous observations (frames) are included during the pose estimation step and no loop closure is performed. Thus, VO is prone to drift over time and does not provide a globally consistent path. Additionally, VO only provides a pose estimate to the initial pose and not to the workspace. Compared to VO, Simultaneous Localization and Mapping (SLAM) introduces the notion of a global map and therefore can ensure consistency by performing loop closure. However, map building, storing and loop-closure detection can be computationally and memory demanding.

In this work, we propose a lightweight, cost effective motion capture system based on a monocular, visual-inertial SLAM system that tightly fuses inertial measurements and observations of artificial visual landmarks, also known as ''fiducials`` which constitute the map. By using artificial landmarks that provide rich information, the estimation, mapping and loop closure effort is minimized. In this implementation, we use AprilTags \cite{olson2011apriltag} as our fiducials. Since these tags also provide a unique identification number, they can be robustly tracked and estimated in the applied Extended Kalman Filter (EKF). Additionally, loop closure is handled implicitly and no additional loop closure detection step is required. A single observation of a tag is sufficient to estimate the relative transformation between tag and robot.

Most tag based localization systems use the relative pose estimates from the tag observations. In this work, we chose a tightly coupled approach where the \emph{corner detections} are used as observations, forming a holistic sensor fusion algorithm. Hence, the system can work with very few tags and observations while still remaining accurate. This reduces the map size of the estimation problem and lowers computational demands. Therefore, the complexity of the proposed system is much lower than common SLAM approaches while still providing accurate, globally consistent estimates. By also including inertial measurements, robust performance during fiducial occlusion, motion blur from fast motions and changing lighting conditions is ensured. The approach requires to artificially prepare the workspace but also provides relative pose information within the workspace. Hence, instead of an alternative for SLAM solutions, we see the developed system as a lightweight tool that can be used as an inexpensive outdoor-capable motion capture system, for verification of other state estimation systems or for absolute localization in a given workspace.

\subsection{Related Work} 
Many existing fiducial-based localization systems are targeted at augmented reality or were designed to be used with cameras only. Hence, many systems use vision data only (e.g. \cite{zickler2010ssl,fiala2004artag,olson2011apriltag,faesslermonocular,breitenmoser11}). These systems have two major drawbacks over the presented system. Firstly, they fail to provide any estimate during occlusion or motion blur. Secondly, linear velocities and body rates can only be computed based on the position and orientation estimates and are thus highly quantized. While this might be negligible for virtual reality applications, it can cause issues when closing a control loop over these estimates. To mitigate these issues, a motion model can be assumed \cite{lim09}. However, this makes the approach specific to the implemented motion model.

The motion estimation and map building elements of the presented approach are closely related to monocular, visual-inertial SLAM, which has proven to be very effective \cite{mourikis2007multi, kelly2011visual, jones2011visual, nutzi2011fusion, li2013high}. The difference between our approach and fiducial-free visual-inertial SLAM solutions is that we tightly integrate artificial landmarks that result in highly robust and unique features in image space. As a result, our landmarks can be robustly redetected and their detections are almost outlier free which increases the robustness of the approach. Additionally, each landmark has a 6-DoF pose (position and orientation) rather than only 3-DoF as in commonly used point landmarks. Furthermore, since a single measurements fully constrains the 6-DoF relative pose, a single landmark is sufficient for estimating the pose. This also allows for a simple yet accurate landmark initialization which usually is a problem in monocular SLAM approaches \cite{sola2012impact}. Furthermore, pose landmarks allow for aligning the map and the estimates to a given frame in the workspace and thus, the system can provide an absolute localization in the workspace which can be crucial for tasks that assume a prepared environment.

While both, fiducial-based localization and SLAM are well studied problems, not many approaches exist that combine both. One approach where fiducials are combined with SLAM is presented in \cite{maidi09}. However, the inertial measurements are not used to estimate velocities but only used for a fall-back pose estimation if all fiducials are occluded. Another similar system as the one presented in this work has been described in \cite{foxlin03}. Since this work is part of the development of a commercial product (InterSense IS-1200) the authors remain relatively vague about their sensor fusion algorithm as well as the achievable performance of their system. Furthermore, dedicated hardware is required which poses additional costs for the user and contradicts the goals of this project to provide a cost-efficient, open source framework.
A third visual-inertial, fiducial based localization system is presented in \cite{you2001fusion}. While also here inertial measurements and visual data are fused in an EKF, the approach does not include measurements from an accelerometer which can be helpful during fast linear motions and can provide a notion of gravity. Additionally, it is assumed that the poses of the tags are perfectly known in a workspace frame. Therefore, one can only place the tags in known configuration and imperfect calibration will lead to an inconsistent map.
In \cite{bryson2007building} a fiducial-based SLAM approach is presented. However, here the fiducials are only represented as point features and thus only the 3D positions of the fiducials are estimated.

\subsection{Contributions}
We present a lightweight motion estimation system based on monocular visual-inertial EKF-SLAM using artificial landmarks. This work tightly couples two proven concepts, SLAM and fiducial-based localization, by using corner observations of 6 DoF landmarks and adapting the corresponding Kalman filter innovation term. This allows for smaller map sizes and leaner estimation. In contrast to existing approaches relying on 6 DoF fiducials, the presented framework processes visual measurements and inertial data in a single estimator which results in consistent data and avoids precalibration and recalibration.
We have developed this tool out of a need for an open source, lightweight, accurate visual-inertial motion capture system. We provide the system as free to use open source software. Since it only relies on standard hardware (an IMU and a camera) and a Robot Operating System (ROS) software interface, both often available on robotic platforms, it can be deployed easily. The source code, the datasets as well as a more detailed technical manual can be found at \url{https://bitbucket.org/adrlab/rcars}, allowing easy integration and full reproducibility of the presented results.

\subsection{Notation and Conventions}
In the following sections, scalars are indicated with small letters (e.g. $f_x$). 
Vectors are indicated with small, bold letters (e.g. $\b{r}$). Matrices are indicated by
non-bold capital letters (e.g. $K$). A capital subscript leading a variable name describes the coordinate frame that the quantity is expressed in.
Position vectors are denoted by $\b{r}$. The trailing subscript describes the direction of the vector from its origin to its goal position (read from left to right), e.g. $_A\b{r}_{QP}$ is a position vector expressed in frame $A$ that points from point $Q$ to point $P$. Quaternions are denoted by $\b{q}$. The trailing subscript denotes the coordinate systems involved in the passive rotation, e.g. $\b{q}_{AB}$ represents the passive rotation from the coordinate system $B$ to the coordinate system $A$. Hence, to rotate a position vector expressed in $B$ to $A$, we would compute $ _A \b{r}_{QP} = \b{q}_{AB} (_B \b{r}_{QP})$.

\section{System Description}
The present localization system consists of two main components, a detector for the fiducials and an EKF for sensor fusion. In a first step, the image acquired by the camera is undistorted. Afterwards, the detector is run on the image which outputs the corner coordinates  as well as a unique identifier number (id) associated with each detected tag. Furthermore, it estimates the relative transformation between each tag and the camera. This estimation is based on an iterative optimization minimizing the reprojection errors between the projected 3D corner points and their detections in image space. In a second step, the EKF uses the information from re-detected tag corners to estimate the robot's state, including pose, linear velocity and body rates. Additionally, the filter continuously estimates the position and orientation of the tags with respect to the camera coordinate frame. When a tag is seen for the first time, its pose is initialized using the relative transformation between the camera and the tag as provided by the detector. After this initialization, the tag pose will be refined within the EKF by using the reprojection errors of its corners in each subsequent re-observation. To ensure consistency, the extrinsic calibration between camera and IMU as well as the additive IMU biases are also included in the filter state.

\subsection{Fiducials}
Over the past years, a large variety of fiducial systems have been developed. Very popular implementations include ARToolKit \cite{kato2000artoolkit}, ARTag \cite{fiala2004artag},  CyberCode \cite{rekimoto2000cybercode} and multiring color fiducials \cite{cho1998multi}. In our implementation, we use AprilTags \cite{olson2011apriltag} which are 2-dimensional, printable bar codes. The reason for this choice was the achievable high accuracy \cite{olson2011apriltag} and  the numerous available detector implementations in C/C++. In our system, we use the detector implemented in cv2cg\footnote{http://code.google.com/p/cv2cg/}. In our evaluations, this implementation has proven to be fast and providing accurate and robust tag detections.

\subsection{Hardware}
The proposed system requires a camera and an IMU. While the transformation between IMU and camera can be estimated online, the camera intrinsics have to be given a-priori.
In our setup, we are using a Skybotix VI-Sensor \cite{nikolic2014synchronized}. This sensor consists of two cameras in a stereo configuration and an IMU. While the sensor is a stereo camera we are only relying on the left camera in this work. The sensor is set up to output images at 20 Hz and IMU data at 200 Hz.

\subsection{Camera Model}
In this project, we assume a pinhole camera model which is applicable to most cameras with common field-of-views. The expected input for the detector and filter is an undistorted image. Therefore, the user is free to choose a distortion model as long as an undistorted image is provided. In the case of the VI-Sensor we are using a radial tangential distortion model.
The pinhole camera model is represented by the overall projection $\pi$ which depends on the camera intrinsics, i.e., the focal lengths, $f_x$ and $f_y$, and the camera's principle point $\b{c} = (c_x, c_y)$. It maps a 3D point $P$ expressed in the camera coordinate frame, $_V\b{r}_{VP}$, to its corresponding pixel coordinates $\b{p} = \pi(_{V}\b{r}_{VP})$.

\subsection{Filter}
In order to fuse the information gained from the observed tags together with the on-board inertial measurement we implement an extended Kalman filter. Relying on appropriate sensor models, this filter uses the inertial measurements in order to propagate the robot's state and performs an update step based on the available tag corner measurements. In the following paragraphs we will explain the sensor models used and derive the required filter equations. For readability, this derivation is carried out for the case of a single tag, but is directly applicable to the case of multiple tags.

\subsubsection{Coordinate Systems}
In our filter setup we assume different coordinate frames. First, we assume an inertial workspace coordinate system $W$. We assume that gravity points in negative z-direction in this frame. Furthermore, we define the IMU coordinate system $B$ and the camera frame $V$. Finally, we define a coordinate system $T$ for each tag which coincides with the geometrical center of the tag and where z is perpendicular to the tag plane.

\subsubsection{Sensor Models}
First, we introduce the sensor model used for the IMU. It assumes Gaussian noise as well as additive bias terms for accelerometer and gyroscope measurements. This can be formulated as follows:
\begin{align}
	\tilde{\b{f}} &= \b{f} + \b{b}_f + \b{w}_f, \label{eq:imu1} \\
	\dot{\b{b}}_f &= \b{w}_{bf}, \\
	\tilde{\b{\omega}} &= \b{\omega} + \b{b}_\omega + \b{w}_\omega, \\
	\dot{\b{b}}_\omega &= \b{w}_{b\omega}, 
	\label{eq:imu4}
\end{align}
where $\tilde{\b{f}}$ and $\tilde{\b{\omega}}$ are the actual measurements of the proper acceleration and rotational rates, $\b{b}_f$ and $\b{b}_\omega$ are the additive bias terms, and all terms of the form $\b{w}_*$ represent continuous white Gaussian noise processes.

In addition to the IMU data, we will also include measurements related to the observed tags. For this measurements we propose a tight coupling by using a corner reprojection based visual model. Given the relative position and attitude of a specific tag with respect to the camera frame, $\b{_Vr_{VT}}$ and $\b{q_{TV}}$, we can compute the position of the $i^\text{th}$ tag corner $\b{_Tr_{TC_i}}$ (fixed to the tag coordinate
frame $\b{T}$) as viewed from the camera:
\begin{align}
	\b{_Vr_{VC_i}} = \b{_Vr_{VT}}+\b{q_{TV}^{-1}}(\b{_Tr_{TC_i}}).
\end{align}
By using the camera projection map $\pi$, we can project the above quantity onto the image plane and derive the corresponding pixel coordinate measurement $\tilde{\b{p}}_i$, where we assume an additive Gaussian noise model
($\b{n}_{p,i} \sim \mathcal{N}(0,\b{R}_p)$):
\begin{align}
	\tilde{\b{p}}_i = \pi(\b{_Vr_{VC_i}}) + \b{n}_{p,i}. \label{eq:corner}
\end{align}

The advantage of the selected visual measurement model is that the noise is modelled directly on the pixel location of the detected corners. While the detector provides an estimation of the tag pose relative to the current camera frame and this could be directly used within the filter, fitting an accurate noise model to this relative pose would have been difficult since the magnitude of the noise strongly depends on the current location and orientation of the tag in the camera frame. In contrast, the noise on the reprojected tag corners is, to a large extent, indifferent with respect to the camera pose and can thus be assumed to be constant and identical for all tags and measurements.

\subsubsection{Filter States and Prediction Model}
The above visual sensor model assumes the knowledge of the tag pose. Instead of using fixed values, which could quickly lead to inconsistencies, we propose to include the pose of the tag into the filter state. Therefore, the filter will be able to refine the tag pose and ensure map consistency.
Employing a robocentric representation of the sensor state and the tag pose, we get the following filter state:
\begin{align}
	\b{x} :=& \left( \b{r}, \b{v}, \b{q}, \b{b}_f, \b{b}_\omega, \b{r}_T, \b{q}_T, \b{r}_V, \b{q}_V \right) \\
	:=& \left( {}_B\b{r}_{WB}, {}_B\b{v}_{B}, \b{q}_{WB}, {}_B\b{b}_f, {}_B\b{b}_\omega, \right. \nonumber \\
	& \left. {}_V\b{r}_{VT}, \b{q}_{TV}, {}_B\b{r}_{BV}, \b{q}_{VB} \right) .
\end{align}
In the above state, $\b{r}$, $\b{v}$, and $\b{q}$ are the robocentric position, velocity, and attitude of the sensor. Furthermore, $\b{r}_T$ and $\b{q}_T$ are used for parametrizing the pose of the tag, while $\b{r}_V$ and $\b{q}_V$ represent the extrinsic calibration between IMU and camera. Computing the total derivatives of the selected state and inserting the IMU model (\ref{eq:imu1})-(\ref{eq:imu4}) yields:
\begin{align}
	\dot{\b{r}} =& -(\tilde{\b{\omega}} - \b{b}_\omega - \b{w}_\omega)^\times \b{r} + \b{v} + \b{w}_r, \label{eq:sde1}\\
	\dot{\b{v}} =& -(\tilde{\b{\omega}} - \b{b}_\omega - \b{w}_\omega)^\times \b{v} \nonumber \\
	& + \tilde{\b{f}} - \b{b}_f - \b{w}_f + \b{q}^{-1}(\b{g}), \\
	\dot{\b{q}} =& -\b{q} (\tilde{\b{\omega}} - \b{b}_\omega - \b{w}_\omega), \\
	\dot{\b{b}}_f =& \ \b{w}_{bf},	~\dot{\b{b}}_\omega = \ \b{w}_{b\omega}, \\
	\dot{\b{r}}_{T} =& -\b{q}_V \left( (\tilde{\b{\omega}} - \b{b}_\omega - \b{w}_\omega)^\times (\b{q}_V^{-1}(\b{r}_T) + \b{r}_V) + \b{v} \right) \nonumber \\
	& + \b{w}_{rt}, \\
	\dot{\b{q}}_{T} =& -(\b{q}_T \otimes \b{q}_V) (\tilde{\b{\omega}} - \b{b}_\omega - \b{w}_\omega + \b{w}_{qt}), \\
	\dot{\b{r}}_{V} =& \ \b{w}_{rv}, ~\dot{\b{q}}_{V} = \ \b{w}_{qv}.
\end{align}
We include additional continuous white Gaussian noise processes $\b{w}_{r}$, $\b{w}_{rt}$, $\b{w}_{qt}$, $\b{w}_{rv}$, and $\b{w}_{qv}$ in order to excite the full filter state and for modeling errors caused by the subsequent discretization of the states. For all white Gaussian noise processes $\b{w}_*$, the corresponding covariance parameters $\b{R}_*$ describe the magnitude of the noise. While most covariance parameters can be chosen by considering the corresponding sensor specifications, some remain as tuning parameters.
Using a simple Euler forward integration scheme a set of discrete time prediction equations can be derived. In order to achieve a minimal and consistent parametrization, the derivatives of the quaternions are expressed in a 3D local angular velocity. This has to be considered during the discretization and when implementing the filter.
Also note that, during the prediction, the IMU-related states $(\b{v},\b{b}_\omega)$ are coupled to the estimated tag pose $(\b{r}_T,\b{q}_T)$ based on the IMU-camera extrinsics estimates $(\b{r}_V,\b{q}_V)$.

\subsubsection{Update Model}
The update step is performed by directly employing the reprojection error as the Kalman filter innovation term. For each tag corner $i$ and based on equation (\ref{eq:corner}) we can define an innovation term $\b{y}_i$:
\begin{align}
	\b{y}_{i} = \tilde{\b{p}}_i - \pi(\b{_Vr_{VC_i}}).
\end{align}
This results in a 8D innovation term for every tag detected in the current camera frame (2D per tag corner). For each newly observed tag the state is augmented by an additional tag pose, i.e. position and attitude. The augmentation uses the estimated
relative pose from the tag tracker in order to initialize the state with a good linearization point. The corresponding covariance matrices are initialized to large values and typically converge very quickly.
Optionally, tags with known absolute location can also be fed to the filter. Especially for datasets with a large number of tags this comes in handy since the EKF does not scale well with increasing state dimension. Above around 20 tag poses in the filter state the prediction step becomes very costly for a single core implementation.

\section{Results}
In order to assess the performance of the proposed approach, we define different test procedures. In a first test, we verify the accuracy of the fiducial pose estimation. In a second test, we then evaluate the accuracy of the motion estimation computed by our EKF. Both tests are verified with ground truth data obtained from a high class external motion capture system. Additionally, two large scale datasets are processed for verifying the accuracy when closing larger loops. A third test evaluates the estimation of the extrinsic calibration. Lastly, we test the applicability of the presented motion estimation within an online closed loop control on a quadruped robot.

\subsection{Datasets}
\begin{figure}[htbp]
    \centering
    \includegraphics[width=\columnwidth]{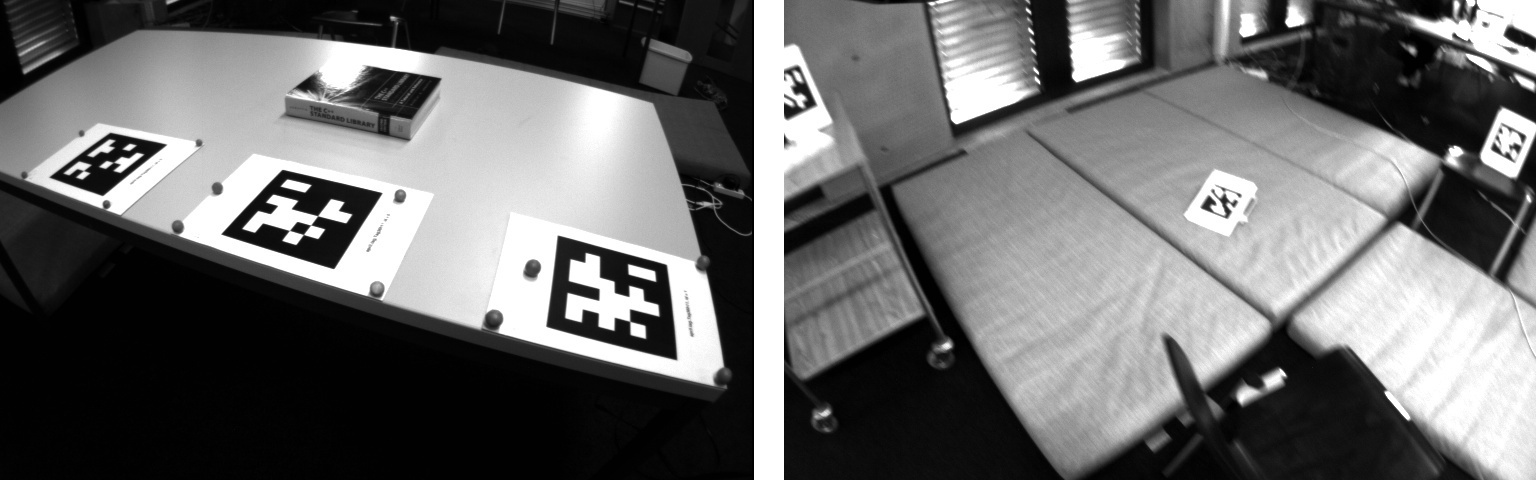}
    \caption{Images extracted from datasets ''table`` (left) and ''dataset\_1`` (right).}
    \label{fig:table}
\end{figure}
In total, we are using five datasets which are all available for download with the source code. The first dataset ''table`` consists of three tags that are placed flat on a table at the same orientation as shown in Figure \ref{fig:table}. The distances between the tags are chosen to be of similar magnitude. 
The second dataset ''dataset\_1`` also contains three tags. This time, we tried to create a challenging dataset, where the tags are sparsely distributed around a larger workspace of about 4x4x4m. Furthermore, the tags are intentionally oriented and located in such a way that the viewing angle for the camera is not ideal and that only a minimal amount of frames contain two neighboring tags at the same time. Additionally, the sensor is moved fast, such that motion blur occurs occasionally. Overall, this increases the level of difficulty in estimating the tags' locations. An on-board image taken by the camera, showing the challenging setup as well as the motion blur is shown in Figure \ref{fig:table}.
For evaluating the extrinsic calibration estimation, we are using a set of datasets all taken within the same workspace to ensure a consistent setup. In these datasets, the sensor is subject to extensive motion in order to properly excite the full filter state and thereby promote the convergence of the estimated extrinsic calibration.
The last two dataset ''cube`` and ''pavillon`` contain round trips on our campus. Both datasets include around 35 tags and span areas of approximately 25x25x6 m. By moving from basement to ground level or from indoors to outdoors, these datasets are subject to changing lighting conditions.
To provide comparable results, no test specific parameter tuning has been performed, i.e. the same parameters are used throughout all tests.

\subsection{Fiducial Estimation Test}
For the verification of the continuous fiducial estimation procedure, we use the ''table`` dataset. In this dataset, manually measuring the offset between the tags is simple. Thus, we can use these measurements as ground truth information and compare it to the estimates of the external motion capture system. This allows us also to evaluate the accuracy of the marker and coordinate system placement during the set up of the external motion capture system. To isolate the fiducial estimation for testing, we are disabling the extrinsic calibration in this test and use the sensor's factory calibration.

\begin{figure}[tbp]
    \centering
    \includegraphics[width=\columnwidth]{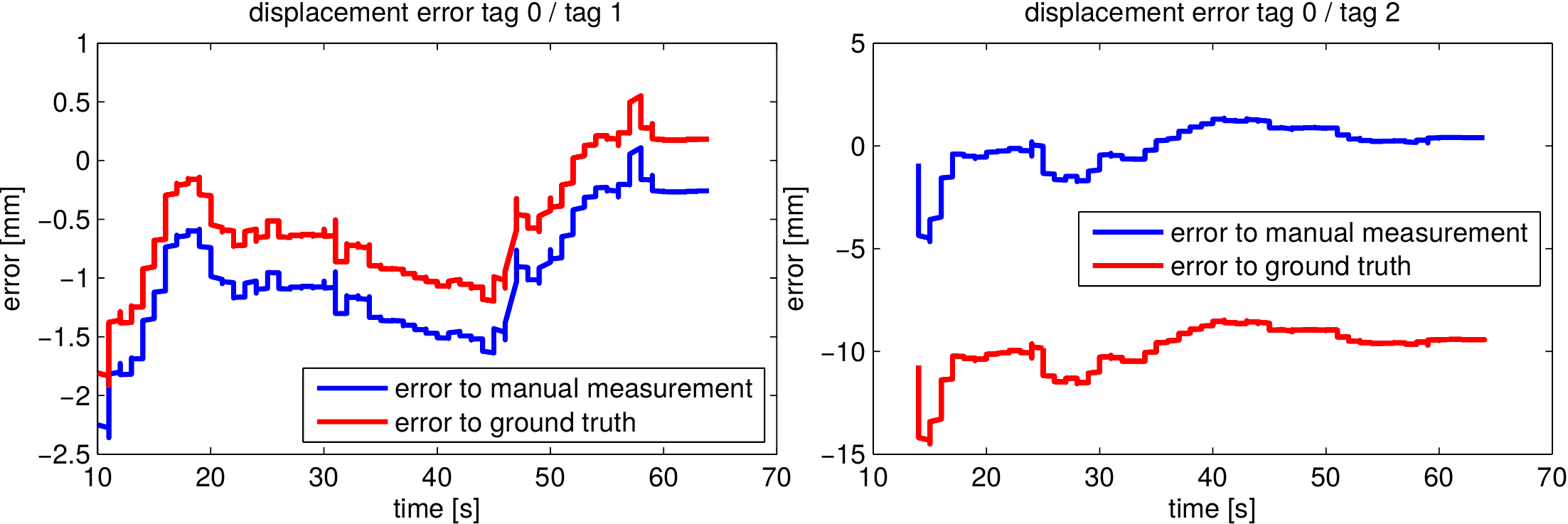}
    \caption{Displacement error between estimated tag positions and reference from manual measurements as well as external motion capture. As can be seen, the error decreases over time, since the tags' positions are iteratively refined by the EKF. Finally, submillimeter accuracy is achieved. The larger offset on the right plot most likely results from inaccurate marker and coordinate system placement in the external motion capture system.}
    \label{fig:pos_error_tag}
\end{figure}

We compare the norm of the relative translation, i.e. the distance between tag 0
and tag 1 as well as between tag 0 and tag 2 with the manual distance measurements. 
This error plots are shown in Figure \ref{fig:pos_error_tag}. 
The plots shows two interesting aspects.
The errors in the beginning of the sequence is quite small. This indicates that the initial guess obtained from the reprojection error optimization on the first frame is fairly accurate. Over time, our EKF then further refines the poses, reaching approximately millimeter accuracy which is of equal magnitude as manual measuring errors.

% TODO Mach doch beide bilder nebeneinander
The figures also shows the error with respect to the external motion capture measurement. Here,
the error is shifted by about 1cm for the translation between tag 0 and tag 2. 
Since a zero mean error curve would be expected, this constant offset most likely results 
from inaccurate marker and reference coordinate system placement.
\begin{figure}[tbp]
    \centering
    \includegraphics[width=0.47\columnwidth]{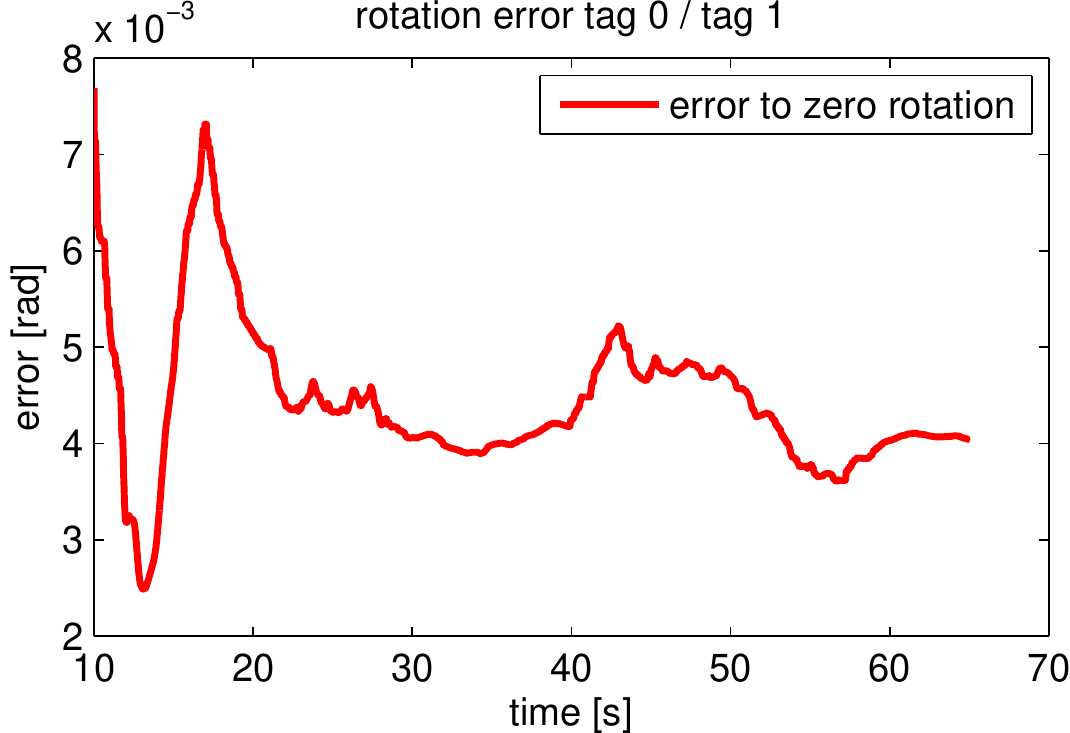}
    \caption{Rotation error between estimated tag positions and zero rotation. The
    error is obtained by converting the relative rotation to angle-axis representation
    of which the angle is plotted. As can be seen, the error decreases over time,
    since the tags' rotations are iteratively refined by the EKF. The error
    starts at around 0.5 degrees and reduce to about 0.2 degrees.}
    \label{fig:rot_error}
\end{figure}
Since the tags in this dataset are placed flat on a table and aligned with the table's
edge, we can also analyze the rotation error of our tag pose estimates. 
To do so, we compute the relative rotation between two tags. We then convert this rotation
to an axis-angle representation and use the angle as our error measurement. Due to the tag alignment, the relative rotation between two tags can be assumed to be identity. This is also confirmed by the external motion capture system up to the fourth decimal of the relative rotation angle. Figure \ref{fig:rot_error} shows the error between estimated rotation
and the identity rotation for the relative rotation between tag 0 and tag 1. 
As can be seen, the error is initially around 0.5 degrees. Through
continuous refinement of the tag poses within the EKF, this error reduces to around
0.2 degrees over time. This error is of same magnitude as printing and measurement 
accuracy.

The experiments described above show the high achievable fiducial estimation accuracy in translation and rotation. 
Furthermore, these results underline that tag pose refinement
significantly reduces displacement and rotational errors present in the single
frame pose estimate used for initialization. This will eventually improve the consistency of the relative tag poses and thus should also improve the robot's pose estimation.

\subsection{Motion Estimation Test}
\subsubsection{Dataset \textit{Table}}
\begin{figure}[tbp]
    \centering
    \includegraphics[width=0.9\columnwidth]{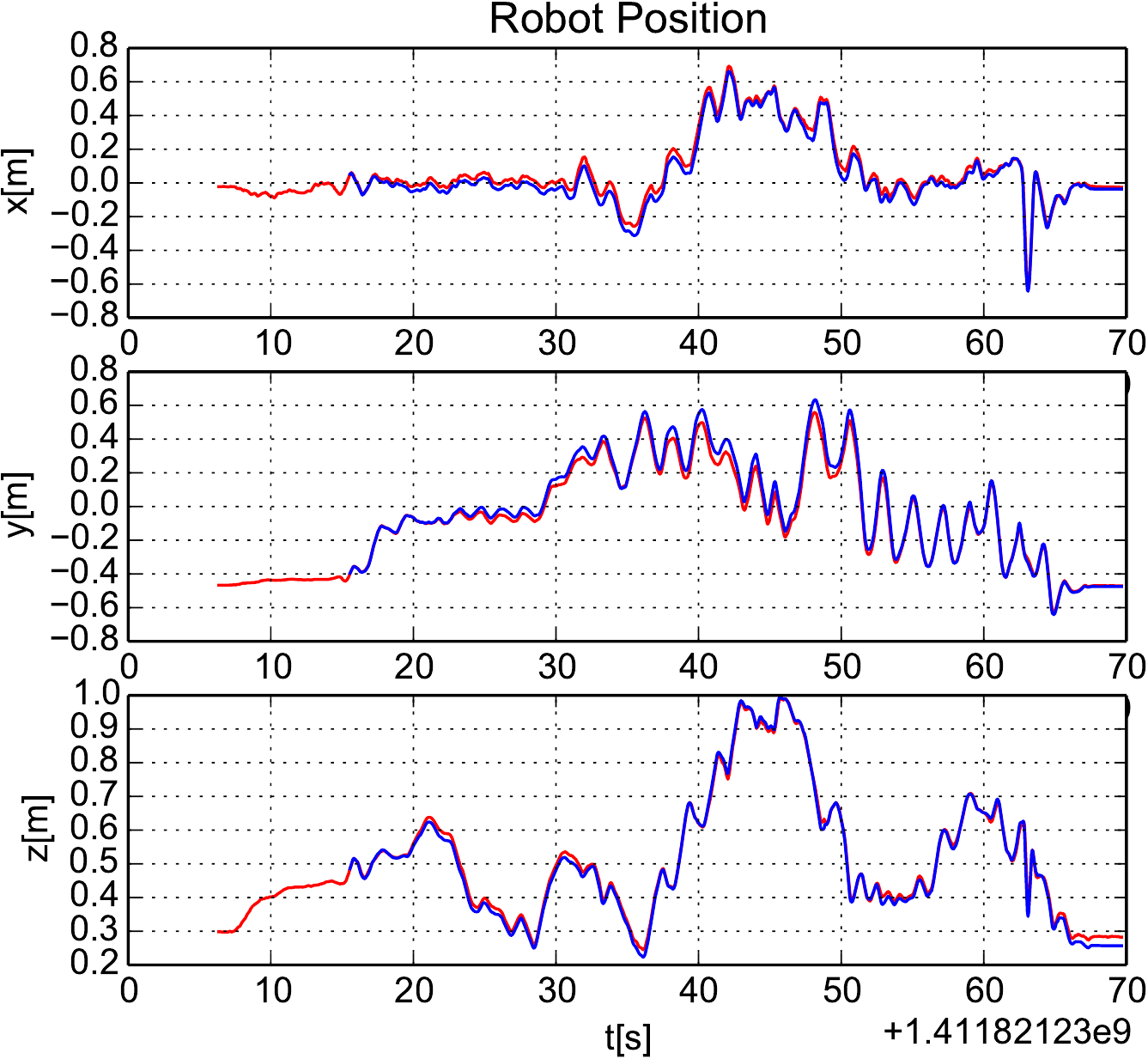}
    \caption{Comparison between estimated robot position (blue) and ground truth position (red) for
    the dataset ''table``. As can be seen, the maximum position offset between both measurements
    lie only within a centimeter scale which is the same magnitude as the achievable measurement accuracy
    in this setup.}
    \label{fig:positionTable}
\end{figure}
To assess the performance of the motion estimation, we use both datasets described
above. The goal of our estimation framework is to localize against our workspace,
where we choose tag 1 as the origin. This choice is arbitrary and one could choose
any tag as a reference defining the workspace location and orientation. Since
our estimator automatically estimates the orientation of the workspace with respect
to gravity, no manual alignment is required.
Figure \ref{fig:positionTable} shows a comparison of the position estimates of the filter and ground
truth data from the external motion capture system for the \textit{table} dataset. This plot nicely illustrates the robust tracking behavior of the system. Even though the reference tag is not detectable at every instance
of the dataset, the estimated fiducials provide a stable reference for the filter to localize against,
such that tracking errors remain a few centimeters.
\begin{figure}[tbp]
    \centering
    \includegraphics[width=0.9\columnwidth]{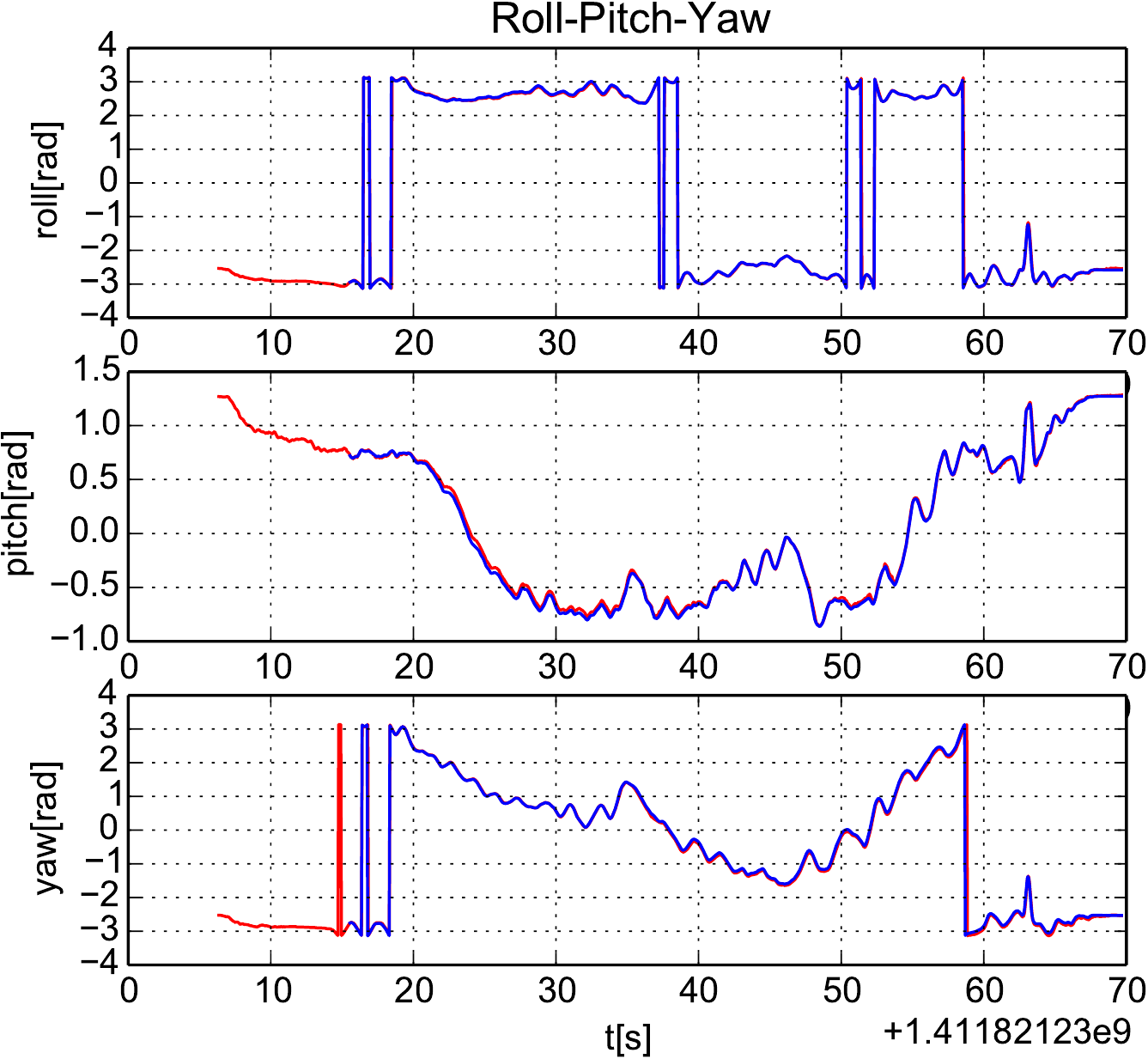}
    \caption{Comparison between estimated robot orientation (blue) and ground truth orientation (red) for
    the dataset ''table``. Due to the wrap-around at +/-$\pi$ the plot is discontinuous. However, since
    quaternions are used for the internal representation of the filter, the output of the filter is smooth.
    As also seen in the position data, estimated and ground truth rotations agree up to measurement uncertainty.}
    \label{fig:orientationTable}
\end{figure}
In Figure \ref{fig:orientationTable} the estimated orientation and ground truth orientation for
the same datasets are compared. Also here, the estimator shows a robust tracking with minimal
deviations. The maximum error observed in pitch direction is about 0.05 rad which corresponds 
to less than 3 degrees. Since the ground truth reference data is a relative pose between the sensor and the 
reference tag computed from the individual poses, the error magnitudes observed above
lie within the measurement accuracy of the ground truth data. While this underlines
the performance of the approach, it does not give any indication about its limits.
Therefore, we tried to push the system to its limits using \textit{dataset\_1} which contains
several artificial challenges as described above.

\subsubsection{Dataset \textit{Dataset\_1}}
\begin{figure}[tbp]
    \centering
    \includegraphics[width=0.9\columnwidth]{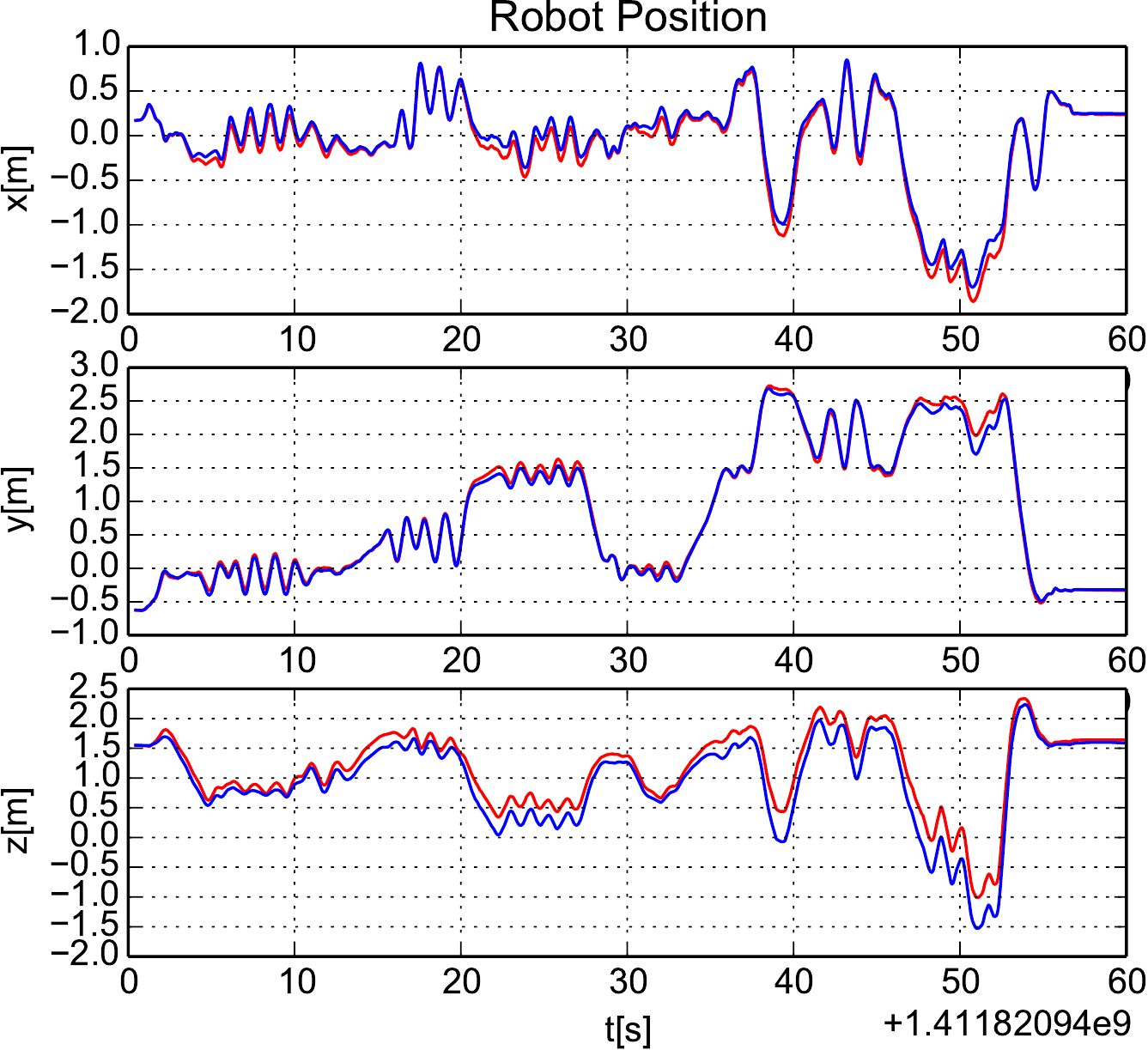}
    \caption{Comparison between estimated robot position (blue) and ground truth position (red) for
    the dataset ''dataset\_1``. This dataset has been made artificially difficult with sparse tag
    coverage and fast motions to show the robustness of the filter. While the estimates diverges
    when only the briefly observed tag on the very left can be used for localization, it converges
    back to the ground truth information when localizing against the other tags again.}
    \label{fig:positionDataset1}
\end{figure}
\begin{figure}[htbp]
    \centering
    \includegraphics[width=0.7\columnwidth]{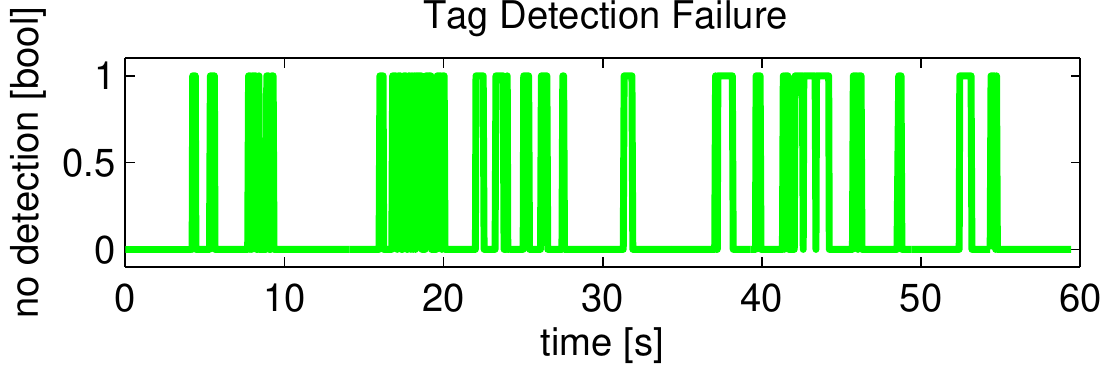}
    \caption{Plot indicating whether one or multiple tags were detected (indicated as 0) or no tag
    was detected (indicated as 1) for \textit{dataset\_1}. Overall, in almost 20\% of all images
    no tag could be detected.}
    \label{fig:tagDetections}
\end{figure}

In this experiment, again the estimated position is compared to ground truth data and the results are
shown in Figure  \ref{fig:positionDataset1}. As the plots show, the position starts to deviate from
ground truth in the last third of the sequence.
While results are not as good as in the \textit{table} dataset, \textit{dataset\_1} can be
seen as a worst-case benchmark scenario. Most of the difficulties for the algorithm are
artificially posed and the performed motion is faster than in many robot applications. Due to
the sparse tag placement and fast motions, the detector was unable to detect any tag in many of the images of the sequence. This
is shown in Figure \ref{fig:tagDetections} where these instances are marked with the value 1. In
total, the filter is provided with inertial measurements only for almost 20\% of the sequence.
Additionally, tag 0 is only seen together with another tag for in total 9 frames. Thus, little
localization information is provided for this tag, leading to a high uncertainty of the tags pose.
Still, it is the only visible tag for about 15\% of the dataset. Thus, the filter is only
provided with uncertain vision information and noisy inertial measurements during these parts. However,
the filter remains stable and is able to converge close to ground truth data again when the
other tags are visible again.

\begin{figure}[htbp]
    \centering
    \includegraphics[width=0.9\columnwidth]{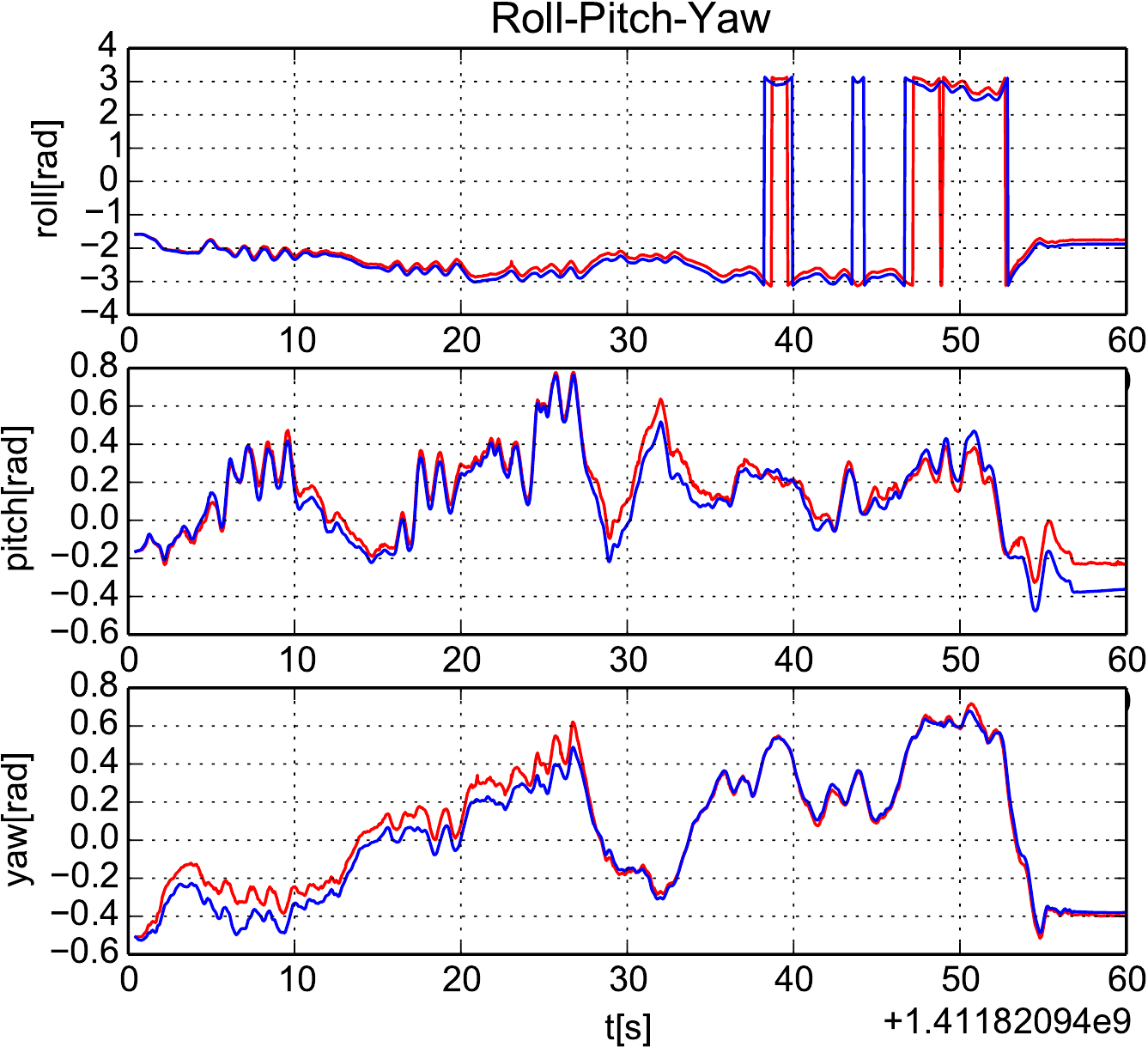}
    \caption{Comparison between estimated robot orientation (blue) and ground truth orientation (red) for
    the dataset ''orientation\_1``. Due to the wrap-around at +/-$\pi$ the plot is discontinuous. However, since
        quaternions are used for the internal representation of the filter, the output of the filter is smooth.}
    \label{fig:orientationDataset1}
\end{figure}
\begin{figure}[htbp]
    \centering
    \includegraphics[width=0.9\columnwidth]{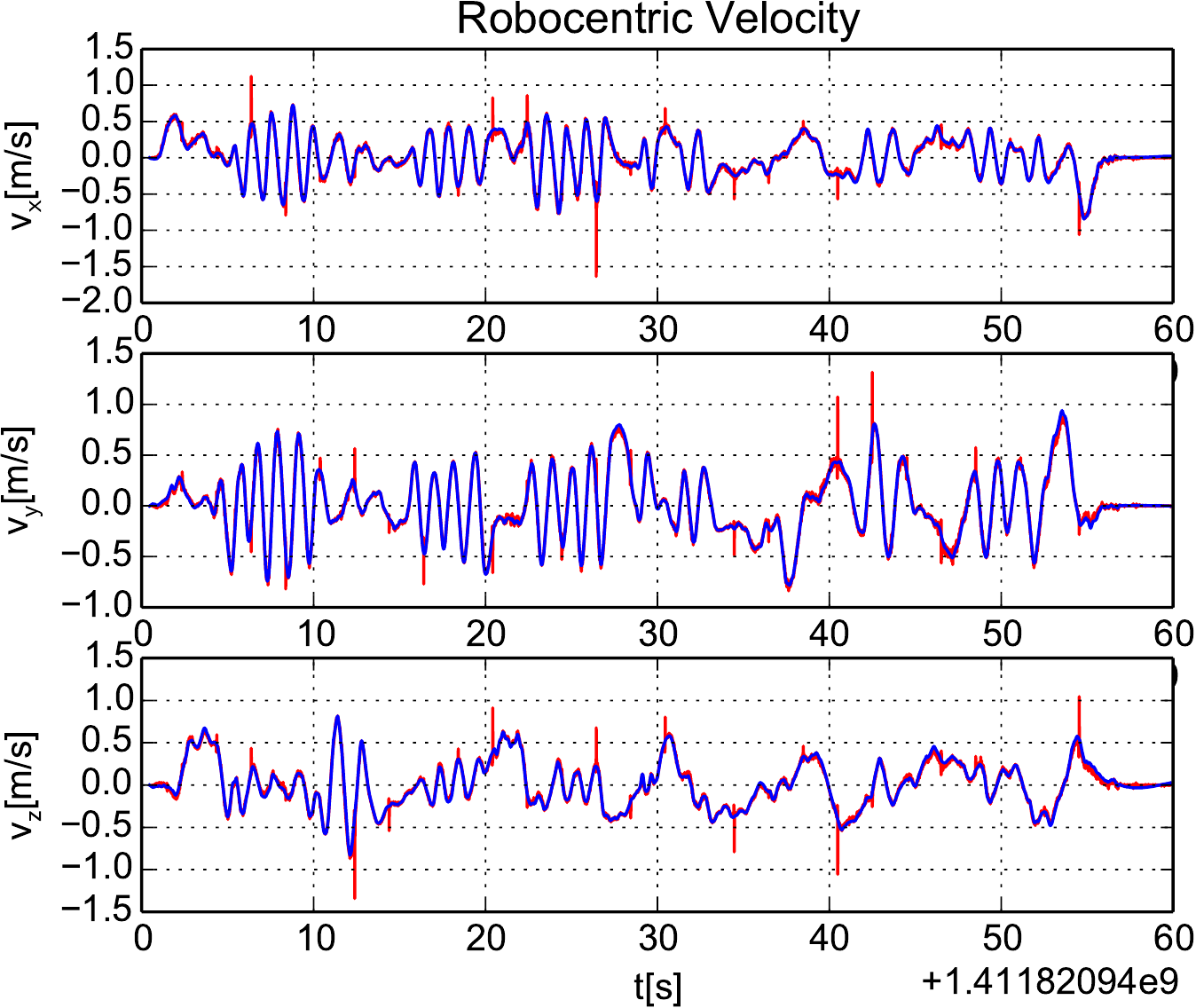}
    \caption{Comparison of linear velocity estimates (blue) and linear velocities calculated by
    using finite differences of the ground truth position data (red). While the estimated velocities
    agree well with the velocities computed from ground truth data, they are virtually outlier
    free. While a high quality external motion capture system is used, this data still shows the limitations
    of finite differences for velocity estimates.}
    \label{fig:linVel}
\end{figure}

Also in the orientation, the effects of sparsely distributed tags combined with
fast motions are visible. Figure \ref{fig:orientationDataset1}
shows the difference between ground truth and estimated orientation for \textit{dataset\_1}. As can be
seen, the orientation estimate is fairly accurate throughout the dataset with a slight deviation
in yaw at the beginning of the trajectory and a small deviation of pitch of about 9 degrees towards the end.
When looking at the linear velocity estimates for this dataset shown in Figure \ref{fig:linVel}, one
can see that the estimates agree well with the velocity data obtained by using finite differences on
the ground truth data. Interestingly, the estimated velocities are virtually outlier free while the
finite differences show occasional peaks. This effect still occurs, even though a high quality motion
capture system has been used. This underlines the limitations of using finite differences for velocity
estimates and encourages the use of inertial data.
\begin{figure}[htbp]
    \centering
    \includegraphics[width=0.9\columnwidth]{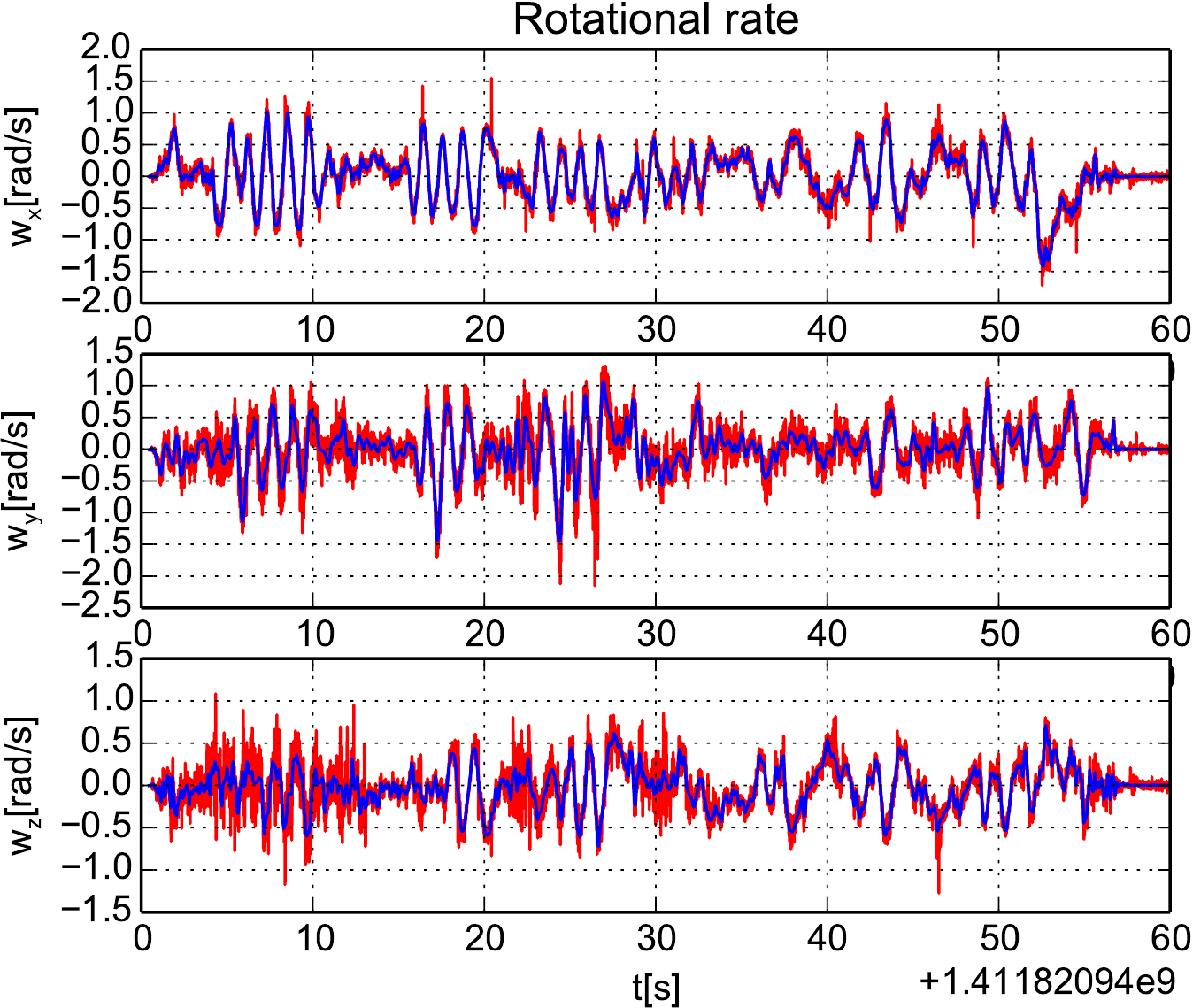}
    \caption{Comparison of rotational velocity estimates (blue) and rotational velocities calculated by
    using finite differences of the ground truth orientation data (red). Like also with the linear velocities
    shown in Figure \ref{fig:linVel}, the estimation matches the ground truth data. Here the significance
    of using inertial measurements for low-noise estimates over finite differences on pose information
    is even more prominent.}
    \label{fig:rotVel}
\end{figure}
This effect is even more pronounced when looking at Figure \ref{fig:rotVel} which shows the rotational
velocity estimates and their counterparts computed using finite differences on the ground truth orientation.
The difference in noise level between both measurements is significant. One reason is that the IMU directly
measures rotational rates using gyroscopes. Furthermore, rotations tend to be more difficult to estimate
for external motion capture systems. This effect gets amplified when differentiating this noisy signal.

\subsubsection{Large scale datasets}
One advantage of the presented approach over a commercial motion capture system is the workspace size. Since our system only relies on paper printed tags, a large workspace can be covered easily. As we did not have a motion capture system available that is capable of covering such a large area, especially not outdoors, we are using loop closure to estimate the accuracy of the approach. For this test, we are using the datasets ''pavillon`` and ''cube`` which both include loop closure sequences. As a quality measure, the reprojection errors as well as the offset between the estimator's predicted tag position and the detector's instantaneous tag measurement are used. Both measures are taken at the first time that we reobserve a tag after the round trip, before updating the estimator.

For the dataset ''cube`` the average reprojection error at loop closure is 56.07 pixels. Taking the detector pose estimate as a reference, the position offset is 0.86 m. One round trip until loop closure is about 70 m long and follows a trail of 36 tags. Therefore, the relative position error is around 1.2 \%. For the dataset ''pavillon`` the average reprojection error at loop closure is 51.01 pixels. Taking the detector pose estimate as a reference, the position offset is 0.38 m. One round trip until loop closure is about 80 m and follows a trail of 33 tags. Therefore, the relative position error is around 0.5 \%. Please note that position errors are calculated using the detector estimate. This estimate cannot be assumed to be a ground truth measurement. Therefore, the accuracy figures above are subject to measurement inaccuracies of the detector and based on one measurement only.

%It can be observed that for both datasets the estimator optimizes all tag poses at loop closure to ensure consistency. Consecutive rounds in the same dataset show that the map is consistent and is further refined. Both these effects can be best seen in the video attachment.
%TODO: ADD LOOP CLOSURE

\subsection{Online Extrinsic Estimation}
In order to assess the quality of the estimated camera-IMU extrinsics, we evaluated the corresponding values after the system was sufficiently excited such that the values could converge. Since no real groundtruth references were available for the extrinsics, we evaluated the repeatability of the obtained estimates. For this we recorded 10 datasets within the same environment while performing similar motions with a total duration around 50-60 seconds. The obtained RMS-values were 1.5 cm for the translational part and 0.0035 rad for the rotational part of the extrinsics. Both values ranging near what can typically be obtained through a dedicated calibration routine.

\subsection{Estimation in Closed Loop Control}
Motion capture systems are increasingly used in closed-loop control. Since latency, noise and outliers can significantly deteriorate the closed-loop behavior of the plant, estimation in the loop is a challenging task. Therefore, we test our system in such an application. For this test, we are using our quadruped robot HyQ on a hydraulically actuated treadmill. The control task is to keep the robot in the center of the treadmill by only regulating the speed of the treadmill, i.e. the robot's walking motion is assumed to be a disturbance. The control system is a cascade of an inner velocity and an outer position control loop. The sensor input to the position control loop are the robot's position and velocity in the workspace.

As can be seen from the accompanying video\footnote{\url{http://youtu.be/Ckf1QAuTKqc}}, the closed loop system is able to stabilize the robot's position on the treadmill while changing the walking speed. The estimate of the absolute position of the robot in the workspace allows us to move the robot to the treadmill center during initialization.

\section{Conclusion and Future Work}
In this paper, we have presented an open-source, visual-inertial state estimation system, that tightly integrates monocular SLAM and fiducial based estimation. By relying on standard hardware already present on most robots the system can be applied cost efficiently. Experiments demonstrate its good accuracy and high robustness, which indicates that it could replace an expensive motion capture systems in applications that do not require sub-millimeter precision or very fast update rates only offered by highly expensive motion capture systems. This has been verified by using the system in a closed-loop control task. Large scale tests have demonstrated long term accuracy, map consistency and loop closure refinement.
Experiments under fast motions and sparse tag coverage of the workspace underline the importance of including inertial measurements compared to fiducial only approaches. Furthermore, the inertial measurements ensure high quality translational and rotational velocity estimates which can outperform these of a commercial system.
Results have shown that good coverage of fiducials is important for a good estimation quality. In the future, we aim at supporting differently sized fiducials such that their size can be optimized for their intended location. Furthermore, we will investigate the combination with natural landmarks in order to further improve the estimation accuracy.

\section*{Acknowledgement}
\footnotesize{
The authors would like to thank Sammy Omari, the Autonomous Systems Lab and
Skybotix for their support with the external motion capture system and the VI-Sensor. Furthermore, the authors would like to thank Manuel Lussi for the support with the treadmill experiments. This research has been funded partially through a Swiss National Science Foundation Professorship award to Jonas Buchli.
}

\bibliographystyle{bibtex/IEEEtran}
\bibliography{bibtex/references}

\end{document}